# TOWARDS VISUAL TYPE THEORY AS MATHEMATICAL TOOL AND MATHEMATICAL USER INTERFACE

LUCIUS SCHOENBAUM

*Louisiana State University*
*Department of Mathematics, Baton Rouge, Louisiana, USA 70803-4918*

Abstract. A *visual type theory* is a cognitive tool that has much in common with language, and may be regarded as an exceptional form of spatial text adjunct. A mathematical visual type theory called *NPM* has been under development that can be viewed as an early-stage project in mathematical knowledge management and mathematical user interface development. We discuss in greater detail the notion of a visual type theory, report on progress towards a usable mathematical visual type theory, and discuss the outlook for future work on this project.

## 1. Introduction

Suppose that you are a working mathematician with some promising new results that you would like to share with other mathematicians in your field. It is exactly for this purpose that the library of mathematical symbols and the language and style of mathematics were originally developed. It is no surprise, then, that this very style is an excellent vehicle for your purposes.

On the other hand, suppose that you are a worker at a company in a data-intensive field, and you have a problem for which it is necessary to utilize the tools of mathematics. For example, you may wish to automate a finitary procedure, solve a problem in software verification, or apply a physical or economic model. The aforementioned traditional language and style of mathematics was not designed for your purposes. The needs of such users for new styles and new forms of expression are well known and have been felt now for many decades. Many mediums of communication and tools for interacting with other members of such fields, as well as with software and computers, have been developed in recent decades. This effort has been enormously fruitful and transformative, for example, from them an entirely new branch of science (namely, computer science) has arisen.

However, for those in any third walk of life, different from either of these, a communication gap exists whose natural tendency is not to shrink, but to further widen with the growth of scientific knowledge. At the present time there is rising interest in mathematics and other STEM areas [Mau], a growing recognition of the growing importance of STEM knowledge, and a parallel rise in awareness of this communication gap, as well as broader challenges for STEM education. These developments have already attracted the interest of scientists in several fields, and has been (in part) a driver of the development of new areas such as mathematical knowledge management [Far] and mathematical user interface development and design [JP2014], fields that intersect and interact with multiple large knowledge domains, particularly mathematics itself and computer science, but

*E-mail address*: `lschoe2@lsu.edu`.

also others such as, e.g., educational and cognitive psychology. In this article, two questions are addressed. First, we address the general question of what theoretical space a tool for mitigating the communication gap named above might occupy. In seeking an answer to this question we introduce the notion of a *visual type theory* in the long section 2, after establishing that such a notion is distinct from existing notions in computer science (in particular programming language theory) and educational psychology and therefore necessitates some kind of new terminology and a somewhat new, developing theoretical framework. Then secondly, in the context of the general body of mathematical knowledge, e.g., the broad fields of analysis, topology, geometry, and algebra, each of which contain numerous specializations, we ask: is it possible to construct a *mathematical* visual type theory? We report that the answer is yes, and that a prototype model is presently in development. In section 3, we describe in outline the prototype system, entitled *NPM*. We conclude our preliminary report with an overview of the outlook of the project in section 4.

## 2. Visual Type Theory: Form and Function

2.1. **Cognitive Model: Comprehension of External Representations, Ordinary Memories.** According to the cognitive model of Schnotz and Bannert [SB2, Sch, SB1], the brain's function of comprehension of external imagery or text involves the creation of a mental model and propositional representation, which are (respectively) depictive and descriptive representations constructed via schema-mediated structure mappings from internal representations that more directly correspond to the given external representations (the imagery, or the text, respectively). The distinction between depictive and descriptive representations is understood in terms of the distinction between *symbols* and *icons* due to Pierce [Pie]. A depictive representation consists of iconic signs associated to the content they represent through common structural features (or "similarity"). A descriptive representation consists instead of symbols having arbitrary structure that are associated to the content they represent by means of convention [Sch]. This model can be viewed as a refinement of the earlier dual coding theory of Paivio [Pai] and the multimedia learning model of Mayer [May]. Such models are used by psychologists to form hypotheses which can subsequently be tested by experiment.

We will adopt this model of visual and text comprehension as our basic framework, and we will refer to the propositional representation and associated mental model generated by an external representation, collectively, as an *ordinary memory* (an idealized but convenient notion). An ordinary memory is distinguished from other memory types, for example spatially organized (as opposed to visual) memories [SB1, p. 616]. In [SB1] the ordinary memory of a graphic (imagery consisting of a graph or chart) is divided into a deep structure and a surface structure. The surface structure is "the outward form of the graphic," and the deep structure is "a semantic construct which expresses the meaning of the graphic" [SB1]. For our purposes, we seek an analogous definition that would apply to cases other than graphics, in particular to descriptive as well as depictive representations. Therefore we define the *sketch component* of an ordinary memory to be whatever serves (pragmatically) as the individual's immediate internal depictive representation of the ordinary memory when prompted. The reader may perhaps wish to view this as the associated communication level representation [GMZ, Sch] of the ordinary memory, but we will stick here to the more colloquial terminology of "sketch component", since our entire argument is quite heuristic and only meant to suggest possible refinements, and since the terminology of [GMZ] does not particularly isolate the depictive portion of the communication level representation as our definition seeks to do. Notice that this definition can indeed be applied equally to external depictive as well as external descriptive representations (as prompts). It is reasonable, and we will assume, that this sketch component or depictive communication level representation is immediately placed in the working memory of the individual upon exposure to an external representation of the associated concept or ordinary memory. Due to the capacity limitations of working memory [SC], this can only be an incomplete part of the total content of the ordinary memory, and we might regard whatever lags in a measure of time-of-recall to belong to the associated *deep component* of the ordinary memory.

We can observe that the sketch component is not well-defined or uniform across space or time (in different individuals, or even in the same individual as time progresses). However, often significant uniformity can be observed in populations. For example, we might expect that a well-known person's face is the sketch component of the descriptive representation given by the person's name (for example, an internal visualization of Bill Gates is the sketch component of `Bill Gates`). Likewise, a steaming cup of coffee could be expected to be the sketch component of the ordinary memory of the depictive representation `coffee`. It is also perhaps noteworthy that the sketch component of a memory can be expected to be more well-defined (i.e., uniform/canonical), in general, than the corresponding deep component.

Images presented alongside associated text, or *text adjuncts*, have been shown to improve memnenic performance, see, for example, [LAC, DSL], and references cited there. In [LAC] a taxonomy of decorational, representational, organizational, interpretational, and transformational text adjuncts is defined, and experimental evidence is presented according to which the effectiveness of text adjuncts for mnemonic and comprehension purposes varies according to the taxonomic type, a point that will be significant for the discussion in section 3.3. In [DSL] evidence is given that other variables can affect the magnitude of such effectiveness, such as setting and period of exposure. In addition they found a significant effect on preservation of learning against decay during the one-week time interval between testing that took place immediately after exposure to a text passage and follow-up testing, due to the presence or absence of a graphic (paired with the text) depicting an illustrative visual metaphor (which they classify as a decorative graphic). Altogether, their study lends support to the view that "extra processing prompted by the presence of the graphics creates a cognitive trade-off for learners" [DSL, p. 202] that can be simplistically described as a short-term cost (the extra energy demanded, e.g., by assimilation of the graphic with the text) that has long-term benefits (e.g., reduced memory decay).

2.2. **Mathematical Memories.** We next turn specifically to mathematical concepts in the setting of the cognitive model of section 2.1. While many studies have been done on integrating pictures into learning of science and mathematics (see, for example, [BSC], [CL, p.13]), the author is not aware of any that focus on mathematical *structures*. As is well known, during the late nineteenth and early twentieth century, mathematics underwent several new developments due to the contributions of many people including Dedekind, Hilbert, Ore, Noether, and many others, which transformed mathematics, in the minds of many, from a numerical science to an abstract, structural science [Cor]. In this section we approach this latter, structural view with a philosophical discussion based largely on personal experience and introspection, in order to draw several conclusions that are crucial to our argument. At the time of publication it is too late to explore further relationships with psychology and mathematical knowledge management here (but see for example [KK] which suggests an approach via the theory of framings). We will also ignore for simplicity the *numeracy* or *numerosity* function of the brain [Deh], as our focus is particularly on the production of ordinary memories of mathematical structures.

Let us try a simple thought experiment. Suppose that an individual tries to think of a group (that is, tries to think of *groups*, the well-known objects defined in any first course in abstract algebra). He or she would normally recall, after a little thought, particular experiences or encounters with groups. He or she might perhaps recall properties groups have (e.g., "all elements are invertible"), or a theorem about groups (e.g., the Jordan-Hölder Theorem). These memories would, depending on the individual, vary widely. This is particularly true for the *depictive* memory of groups—and it is just this which we would like to draw a focus on. While there are *examples* of groups that evoke depictive representations (symmetric groups, dihedral groups, and orthogonal groups, for example), and there are *heuristic images* that help us to see what groups in general are like, there is no telltale visual cue, no "first-order association" of a group with a depictive representation—something that, when visualized, prompts the entire mental model of a group. Depending on the individual, he or she might have an intuition at hand—for example, spatial intuition for Lie groups, or model-theoretic intuition for polish groups, or set-theoretic intuition for finite groups—but the group concept itself is beyond all such intuitions, or rather, it is apparently a puzzling marriage of all these intuitions, and more besides that is apparently beyond an individual's power to clarify. In brief, the group concept can be regarded as a "dark" concept.

Indeed, (whether it is fair or unfair to use the evocative term "dark") many other examples can be given to illustrate that this "dark" quality is in fact commonly encountered in mathematics. While in the example of a group given above it arises due to the formal application of mathematics and logic (the axiomatic method), the same dark quality (the lack of a sketch component in memory) also arises due to the general application of idealized infinite (i.e., nonfinitary or nonterminating) processes in derivations of mathematical formulas and relations, as well as the ubiquitous use of quotient operations in mathematics (see below). Consider, for example, the difference between a standard visual representation of the rational numbers (commonly denoted $Q$) and the real numbers (denoted $R$ or $(-\infty, +\infty)$) as ordered structures. In both cases, the ordering is linear and *dense*: between any two elements in the ordering, there can be found another distinct element. Only in the latter case, however, does the ordering have the additional property of *completeness*, which is perhaps most easily understood in terms of a topological structure assumed to be present in both cases, or alternatively, by way of a construction of the latter ordered set from the former one. This property, however, defies visualization: density and completeness are indistinguishable in an external or internal depiction, except via a heuristic (i.e., a diagram that heuristically indicates some phenomenon that cannot be directly visualized). Other examples demonstrating this phenomenon include space-filling curves, real-valued dimension (fractals), and so-called "pathological" functions (for example, the continuous, nowhere-differentiable function discovered by Weierstrass [GO]), many of which arose in the nineteenth century and drove the earliest efforts to place calculus on firm foundations. Or, to reach for examples from other fields, we can think of trying to visualize manifolds (abstract spaces that are curved, like the surface of the earth) that are not embedded in three-dimensional space. The (perhaps) most well-known example of such a manifold, the so-called Klein bottle, has a very famous and lovely depictive representation, but in general, for higher-dimensional manifolds with arbitrarily chosen topological invariants, no such depictive sketch component is available. The same can be said for structures defined only up to some equivalence class ("only up to isomorphism"), i.e., concepts which are obtained via a quotienting operation applied to existing families of concepts. It is not difficult to find examples in which elements of an isomorphism class—for example, the isomorphism class of the plane $R^2$ up to topological isomorphism (i.e., homeomorphism)—are themselves accessible via a natural depictive representation, and yet there is nevertheless no meaningful or pragmatically useful depictive representation of the entire isomorphism class.

If we set these examples in the context of our cognitive model, we can observe that the ordinary process of creating a mental model and a propositional representation, and an associated sketch component, is confused by the richness and complexity of the concept—in particular the lack of a fixed, canonical, uniform visualization. To couch this point in empirical terms, an experiment could be performed in which mathematicians (or some other chosen population) could be asked to draw pictures of mathematical objects. Our hypothesis, then, would be threefold: first, that there would not be found any uniformity in these sketches, second, in fact in many cases such an experiment might well yield descriptive representations serving as proxies to depictive representations (in other words, there is in that individual's mind no properly depictive representation at all, and the individual has simply "doubled up" the descriptive representation to serve dual purposes, depictive and descriptive), and third, that there would oftentimes be little or no pragmatic utility (for the purpose of problem solving or memnonic value) in the sketches produced. Assuming this is accurate, what is therefore observed is a certain inefficient or wasteful use of the brain's cognitive system: immediate or sketch components of the brain's recall of mathematical concepts is of little use to the user or learner of mathematics.

2.3. **Haptical Memories, Barriers to Entry.** It is unclear just how it is, exactly, that mathematicians construct and maintain memories of mathematical concepts. However, if the author ventures to argue on the basis of his own experience, he would cautiously lean towards the view that mathematicians rely often on memory mechanisms of the brain other than the ordinary memory system. In particular, they often rely on what we will refer to here as "muscle memory" or "haptical memory". Units of such haptical memory are produced in the course of problem solving: passing from a question whose answer is not clear *prima facie* to an answer via logical inference and computation, which may be abridged by prior knowledge or by assumptions and conjectures. (This, we believe, is what mathematicians often refer to simply as "working".) Let it be granted, then, for the sake of argument, that mathematicians, at least in large part, rely for their memories of mathematical concepts on a posited haptical memory system that is broadly available in the human population. This haptical memory system is assumed to interact, via structure mappings, with the ordinary memory comprehension system, hence it influences the construction of the propositional representation and—perhaps directly, or perhaps indirectly—the mental model of a concept as well.

Haptical memory tactics can be very powerful. However, while an ordinary memory can be formed in an instant, a haptical memory takes several minutes at the very least, and as noted in [SK] places significant cognitive demands on the brain. This matters little in many situations, but with the sheer number of abstractions present in modern mathematics, and the increasing demands for mathematical knowledge in fields outside of mathematics, the acquisition of mathematics through haptical memory tactics is increasingly impractical in many settings. Moreover, mathematical information, once acquired, still must be managed and maintained, activities for which the brain demands energy (due to being so-called biologically secondary knowledge [SK, p.474]). This establishes high barriers to entry at the boundary between those who are willing and able to invest significant resources to acquire, manage, and maintain haptical memories of mathematical concepts, and those who are not.

Cognitive load theory [SC, SK] poses many potential strategies for confronting this conundrum. The findings in [DSL] noted in section 2.1 lend partial support to an approach focusing on the use of visual adjuncts to the normal descriptive representations of mathematical structures (i.e., the rigorous development and study of mathematical structures, written down in the language of logic). This, by itself, would not seem to be a new idea, indeed modern math and science textbooks are rife with illustrations of mathematical concepts intended to facilitate learning and boost concept retention, and there has been no lack of reflection on and criticism of the famously influential "Bourbaki style". However, in light of the observation in section 2.2 that mathematical concepts are dark, we can focus attention in particular on the sketch component of memory for mathematical concepts, which is, as observed there, (depending on the individual) missing and/or dysfunctional (serving little or no practical purpose). This begs the question: what if this particular dysfunctional part of the full ordinary memory was replaced by some sort of (in the language of [LAC]) transformational representation? The findings in [DSL] indicate that through such a strategy the problems with the significant time demands and rapid decay of haptical memories might be mitigated.

2.4. **Visual Type Theory: Definition.** A *visual type theory* is a system of depictive representations created in the course of implementation of the strategy suggested at the end of the previous section. Such a system must have several properties in order to properly be considered a visual type theory:

(1) It should depictively reflect logical or natural relationships between the objects being depicted.
(2) It should be *almost regular* in the sense that it should be regular in its prescribed use as a general principle, however it should tolerate irregularity when such an irregularity compellingly suggests itself.
(3) It should be *generative* in the sense that there is a natural way to generate new representations based on the patterns of existing representations.
(4) It should possess the property of *density*: the informational content [Sch, p.103] of each representation of the system should be high in proportion to its visual complexity. For this it might rely on *absence loading* (section 3.3).

A visual type theory can, in theory, function as a communication medium as language does. It differs from language, however, by consisting of depictive, not descriptive representations. Thus we arrive at the "visual" part of the term visual type theory. We use the terminology "type theory" because in the visual type theory of section 3 the representations are very similar to types (for variables) written in a "visual" style of notation, and also because calling it a form of type theory reminds us to distinguish it from language while preserving the idea that such a system, like language, is nevertheless (almost-) regular and generative. Visual type theory provides a system of representations that could be described as "anchors" or "visual anchors" that are particularly useful in situations where concepts are otherwise highly abstract (or, in the language of section 2.2, dark). Through the use of a visual type theory an individual acquires memory tactics in addition to the haptical tactics of problems solving and computation, rote associative memory building, etc.

Now that we have defined visual type theory, a word of caution may be in order. No memory is as simple as a sketch component (section 2.1), taken by itself, would indicate. Beyond the sketch memory there may exist an effectively limitless complex of information. In practice, visual type theory can do no more than provide a user with integrated systems of well-chosen sketch memories of concepts. It leaves to the user the job of filling in deeper memories, and acquiring other abilities a user might need that the visual type theory is possibly unable to provide alone. This is true in particular for a mathematical visual type theory and thus, in this case we caution the reader that visual type theory is not (and is not advertised to be) a "Royal Road to geometry," and does not disprove the famous ancient dictum of Euclid. Sketch memories are simplistic and can contain inaccuracies that only deep memories can correct. Sketch memories are accompanied by the hazard of false assurances, overhasty opinions, and misguided actions. However, in light of the observation of section 2.2 that abstract mathematical concepts often lack a sketch component (creating a vacuum that, as argued in section 1 and section 2.2, can also impede mathematical cognition and comprehension), a trade-off is expected between which, perhaps, a balance can be struck.

2.5. **Visual Type Theory as Language.** To conclude this section, we now ask the question: Is a visual type theory a form of language? If so, just what kind of language is it? As a device with a formal definition and prescribed rules of use, a visual type theory (if perfectly regular) may be viewed as being something like a formal language. As a device with a pattern of growth and an internally consistent set of heuristic (though not unbending) principles, it may be viewed as being something like a natural language. Visual type theories also have many properties in common with both types of language: their purpose, like that of all languages, is to express thoughts, display relationships quickly and memorably, to facilitate the possession, understanding, and management of complex and abstract ideas, to port well through different environments, and, potentially, to serve as a method of communication between users who share the necessary prior knowledge. But, as we have argued in section 2.4, visual type theory technically cannot be considered a language at all: it is not a mapping between syntax and semantics, or in other words, a cognitive tool whose elements travel via the descriptive channels of the ordinary memory system—instead, unlike in a language, they travel via the depictive channels. However, with so many commonalities with existing language notions (and because the use of generally familiar vocabulary to discuss visual type theory is, by all indication, inevitable), it might be wise to bend the notion of language slightly by referring to visual type theory as a *language paradigm*. As such a paradigm, it is distinct from any of the major paradigms that currently exist in computer science: it is not a formal language, nor a compilable programming language, nor a specification language, nor a natural language. From this observation, we can draw several closely related and important corollaries. First, we have seen that visual type theories are not developed as a *language* would be, in order to replace existing mathematical language—neither in whole nor in part. It is not the case that the use of visual type theory would impinge in any way upon mathematical practice if it is used as prescribed here. Second, a visual type theory is not a universal formal system for expressing mathematics, for it operates within an entirely different language paradigm. Rather, it is a supplementary system that runs in parallel with formal languages, whether locally or globally defined. In other words, the depictive character of visual type theory provides it with the property of *independence*: the mathematical formalization a user chooses can be switched out easily while the visual type theory, say, $T$, stays in place, just as $T$ can also itself be removed or added, turned off or turned on at any moment, without affecting the rigor expressed by formal syntax. These formal languages can be of virtually any kind—past, present, or future. This suggests particular uses for visual type theory in mathematical knowledge management, in its role as custodian of past and present mathematical documents and for confronting the problems of informality and logical heterogeneity [RK].

It may be noteworthy that an attempt to change this prescribed role—to use a visual type theory $T$ as a basis for an imagined "visual" formalization of mathematics, for example—regardless of success or failure, would take away this property of independence, and force $T$ to cease being a visual type theory, and to become instead a "visual" formal language.[1] The independence property removes the most stringent pressures that a language can be subject to: the demands of rigor, compilability, security, searchability, etc. In contrast, these pressures fall directly on formal languages and programming languages. Such languages must cater to the needs of machines first, and human beings, second. In contrast, visual type theories, like natural languages, occupy a space in which this priority is exactly reversed.

[1] Many such languages (and similar tools) exist, cf. [Mye].

## 3. Progress Report: A Mathematical Visual Type Theory

3.1. **Overview, Principles of Design.** Now that we have explained at length the simple idea of setting up a visual type theory, we can discuss such a visual type theory for mathematics that is currently under development, called NPM. (The abbreviation NPM originally stood for "not *Principia Mathematica*," but we will not dwell on this.) A system at minimum requires a supply of depictive glyphs (which are intended to serve as canonical sketch components for ordinary memories of mathematical abstractions, cf. section 2.2), rules that govern their use, and tools that make their use in the real world possible. More concretely, we have:

(1) a set of glyphs called the *NPM glyphs*. (more concretely, a scalable font, the *NPM font*).
(2) the grammatical rules governing the use of glyphs (Briefly, the *NPM grammar*),
(3) the "Principles of Design" of the NPM system. (Guidelines for the ongoing development of NPM.)
(4) the correspondence between glyphs and mathematical concepts (the *NPM meaning map* or *interpretation*), and
(5) supporting software/technology (see section 4).

By the Principles of Design we mean to include both the informal or heuristic relationships between mathematical concepts and visual elements of NPM glyphs, and the underlying motives and principles of visual type theory. We also include the following basic principles (hereafter: Principles of Design #'s 1, 2, 3, and 4), which lay at the heart of NPM. They have no relationship with the general notion of visual type theory explained above.

#1. NPM should aspire to be as beautiful as the mathematics it expresses. It should "fit in" with the notation and methods of ordinary mathematical practice.
#2. NPM should be both printable and hand-writable (with a pen or similar stylus).
#3. NPM should have a many-one meaning map between the logical space and conceptual space of mathematics, and be as close to a one-to-one map as possible.
#4. NPM should aspire to be *universal*: to provide one internally consistent set of glyphs for all of mathematics (or more precisely, all reasonably well-established knowledge in all mathematical areas).

Principle of Design #1 and #2 impact various design decisions in the development of NPM glyphs. Principle of Design #2 primarily aims, without specifying a reason, to preserve the traditional bond between haptical experience and mathematical memory. Principle of Design #1, together with the density requirement of visual type theory (section 2.4), suggests that the design be simple, with a minimum of flourishes. This conclusion is supported by studies on text adjuncts which find that learning and memory outcomes may decrease due to interference effects [SB2] and in accordance with a so-called coherence effect or coherence principle [MM, p. 95]. The glyphs should also strive to conform to typographical design standards and principles. For example, it should aim to "induce a state of energetic repose, which is the ideal condition for reading" [Bri, p. 24].

Principle of Design #3 has two parts. First, it says that there should be no overloading of NPM glyphs. Just like the syntax of a strictly typed formal computing language, every NPM glyph should have a definite, fixed meaning, with nothing left to context to interpret. Second, it says that as much as possible, the sketch memory of a mathematical concept provided by NPM should be like a fingerprint: it should be a unique identifier. There are many cases in which this specification cannot be fulfilled. Consider notions that exhibit so-called "cryptomorphism," or nontrivial equivalences between several different definitions. The concept of a matroid, the concept of a lattice, or the concept of a topos are just a few examples of cryptomorphic notions. It is usually sensible to express such cryptomorphism in NPM, rather than choosing one notion and expressing it visually, ignoring the others. Principle of Design #3 tells us, however, to select one of these glyphs and give it precedence over the others. Principle of Design #3 provides a counterweight to the common experience of encountering overloaded notation in mathematics, and disagreement between the notations and terminology used by different authors. It is notable that without changing a single letter or a single word in the canon of mathematical terminology and notation, we obtain in this way a tool for disambiguation that is always near at hand, yet unobtrusive because it is depictive and thus does not share the same cognitive channel as the descriptive symbolism. We note, echoing section 2.5, that this benefit to the use of NPM can be enjoyed without explicitly writing or reading NPM glyphs. We also note that, although NPM commits itself to disambiguation just as formal languages and programming languages do, it does so in response to the needs of a reasoning human being maintaining a large, abstract, technical language, not those of a mechanical parser.

Principle of Design #4 states that NPM should be developed to serve as a tool on *the broadest possible base*, where "base" refers to the potential users of the glyphs and grammar as a cognitive tool—even though, as stated in the introduction, a major goal and motivation of NPM is to increase understanding and lower barriers to entry to mathematics, particularly for those in career paths or life situations which do not permit them to pursue focused, rigorous mathematical studies. Although this may seem like a point of ambition for the NPM project, in the author's view a mathematician would in fact consider it to be the only path forward, given the deep, inescapable connections between mathematical knowledge in different fields. Any attempt to draw a boundary around the parts of mathematics NPM does and does not attempt to assimilate into its linguistic pattern would inevitably be disrupted, due to the natural course that mathematics itself tends to run.

Before taking steps to finalize a mathematical visual type theory, it must be checked whether such a construction could really exist. In particular, it must be determined if it can in fact be universally applicable across all (or at least, some significant majority) of mathematical specializations. Due to the minute, technical details involved, this task cannot be considered completed until a real mathematical visual type theory has been found meeting the specifications to an extent that convincingly indicates that the project can be continued all the way to completion. Until then, it is still not determined whether a visual type theory could exist which holds, in one grammar and one glyph set, all the concepts of mathematics at once, or whether any attempt to create such a system would inevitably burst at the seams, overwhelmed by unexpected, accidental conflicts between patterns inspired by different background knowledge areas. An extensive effort to find an answer to this question has recently been concluded, and it has been confirmed that a visual type theory, conforming to all the above-named specifications, is indeed theoretically possible. (However, difficult challenges of a mathematical nature still remain, as will be discussed further in section 4.) We now describe this system in outline.

3.2. **Grammar.** NPM (viewed as a language) has a large vocabulary of glyphs, but a simple grammar. NPM glyphs are primarily used as identifiers and as modifiers on function and relation symbols. They may appear under arrows, under relations, or standing alone. These are the only prescribed uses of NPM glyphs, though they may also theoretically appear anywhere else they are determined to be useful.

The decoration of mathematical calculations with NPM glyphs performs several functions. It distinguishes calculations in one setting from calculations in a different setting, when the two are formally similar. It also allows users to assemble objects and notions into coherent conceptual wholes, or in other words, to build new abstractions out of preexisting abstractions in a convenient and regular way. Often, these new wholes take the form of categories, or sometimes (depending on the setting) 2-categories (more rarely, even higher categories can appear). It is convenient to express such categories using NPM for several reasons. Such categories are abundant throughout mathematics and can be difficult to manage due to their number. In a given mathematical specialization it is not uncommon for many dozens of categories to arise, even considering only the basic prerequisite knowledge. Categories are also particularly susceptible to the property of having a complicated definition in spite of playing a very basic role in some derived setting or field of investigation. NPM assists in managing such levels of suppressed complexity, creating room to focus on the most relevant details in a given setting. NPM is good at encapsulating complexities, allowing them to be stored neatly, extracted when needed, and stored again when the need is met. Glyphs may be combined in natural ways, providing "ramps" leading from basic ideas to more complex concepts. An NPM expression of a subtle and complex mathematical notion can be a helpful guidepost, and a landmark in the memory of the user that can easily be recalled later, even sometimes used (to one's delight) to reconstruct a forgotten definition. It can also be helpful in drawing attention to relationships between concepts. This is particularly helpful when (for whatever reason) terminology and notation cannot perform the same role, as is sometimes the case. Nonetheless, as the reader can imagine, NPM does not do everything well. Statements which are "decorated" with NPM glyphs can become heavy, and this can become counterproductive and wasteful. This phenomenon reinforces the principle (section 2.5) that NPM might remain in its role as a visual type theory as opposed to being viewed as an extension of mathematical formalism. In this role its explicit appearance in mathematical calculations and statements would, a priori, be optional and left to the user's discretion.

3.3. **Glyphs.** Finally, we address the question of the appearance and design of the glyphs themselves. Although glyphs of NPM bears some resemblance to other sign systems—most notably perhaps, Chinese Han characters and chemical nomenclature—they rely for much of their power on a tool called *absence loading*, which we now define, and which is not a part of the prior knowledge of either of these sign systems.

The space in which a glyph is written or imprinted is called the *body*, or *bounding box*. Absence loading is the association of a spatial region in the bounding box (a region anywhere in the bounding box that may have any size or shape) to a logical constraint $\Phi$. Once this region is assigned, information concerning $\Phi$ may then be associated, via convention, to hand-drawn or printed lines which are placed in that region. In particular, an empty or vacant region (i.e., the absence of any line or mark of any kind) indicates that either nothing is assumed about $\Phi$, nothing is known about $\Phi$, or that information about the status of $\Phi$ is not relevant to the context where the glyph is drawn or printed. (It is left to context to determine which of these cases applies.)

Here is an example. Consider two sets $A$ and $B$ (in some fixed universe $U$). We can construct a simple glyph using absence loading that would tell us whether something, say $x$, is in the set $A$, and whether it is in the set $B$. Suppose that we take the symbol — as a guide for the eye, and "load" or "charge" the two regions below the horizontal line — with information about membership in $A$ and $B$, respectively:

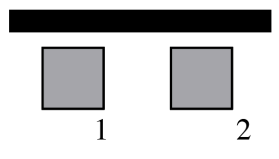

*region* 1: $x \in A$
*region* 2: $x \in B$

Now let us assume that a dot in the loaded regions shall denote: positive, certain membership in $A$, and positive, certain membership in $B$, respectively. Then we can see that our entire language consists of the symbols $\overline{\quad\quad}$, denoting the general universe of sets and elements $U$, $\overline{\bullet\quad}$, denoting the set of elements of $A$, $\overline{\quad\bullet}$, denoting the set of elements of $B$, and $\overline{\bullet\ \bullet}$, denoting the set of elements in both $A$ and $B$. If we close off the grammar and glyph system here, we can see that we have the power to denote only the sets $A$, $B$, and $A \cap B$, along with the universe $U$. However, we can fill the loaded regions with any symbol we wish. Next, suppose we place, say, an open circle $\circ$ in the regions to denote positive, certain *non*-membership in $A$ and $B$, respectively. Then we generate the usual four disjoint classes: $B - A$, $A - B$, $A \cap B$ and $U - A - B$ via the symbols $\overline{\circ\ \bullet}$, $\overline{\bullet\ \circ}$, $\overline{\bullet\ \bullet}$, and $\overline{\circ\ \circ}$. Notice the additional flexibility: we can, using the same symbol system, leave the region assigned to $A$ or $B$ absent, leaving us with a smaller set of symbols germane to that setting, for example, $\overline{\quad\quad}$, $\overline{\bullet\quad}$, $\overline{\circ\quad}$ can serve in any setting in which we are not interested in $B$, only in the set $A$. Hence if we were in a setting in which we wish to ignore some information, we could continue use the glyphs we have defined rather than switching to another set of glyphs. This is achieved by making use of natural visual processing mechanisms: upon exposure to visual information, relative locations in memory are maintained, but visual information considered irrelevant is filtered out in favor of visual information that is the focus of attention. For this reason, the notation is interpreted as a depictive representation, not descriptive one. Certainly, in this notation the amount of depictive power obtained is rather slight, but it's not so bad for a line, two dots, and two circles. What we really mean by this, of course, is that the notation has the property of density that was defined in section 2.4. This density is more noticable if one compares an absence loading notation to other depictive representations of the situation. Consider the most common depictive representation, a Venn diagram:

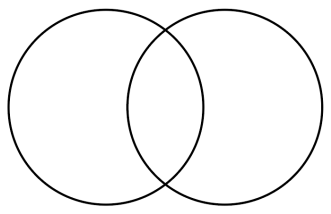

While we will not draw them, the Venn diagram above suggests other possible "visual" notations for the two sets and their subsets. Note that while it is also possible to *extend* such a Venn-diagram-inspired notation to denote three sets, already in the case of four sets, Venn notation is stretched to its breaking point and has practical value only in exceptional cases where some of the intersections are empty. This is not so for the notation using absence loading: given three sets $A, B, C$, we extend the line to produce an immediately evident *extension* of the sign system:

$$\overline{\bullet\ \circ\ \bullet}$$

Given more sets $A_1, \ldots, A_7$, we can continue further: $\overline{\bullet\ \circ\ \bullet\ \bullet\ \circ\ \circ\ \bullet}$, etc. This, for example, would generate a language of $3^7$ or 2187 unique[2] glyphs, or something close to the number of Chinese Han characters in frequent use [L]. This extension of the notation naturally suggests itself because the use of spatial regions of the bounding box naturally induces visual patterns. The corresponding descriptive representation (using formal logical symbols) can easily be derived from the depictive representation, and vice versa. Although this set of examples has little practical value, absence loading can provide a potentially useful notation alternative to descriptive notation in any setting in which frequently recurring conditions $\Phi, \Psi$, etc. arise in highly variable combinations. Such settings are found in many technical and scientific knowledge areas.

[2]Unique up to equivalence defined by the Gestalt laws.

Now we come to apply absence loading to the generation of mathematical glyphs, and the design choices specifically made in the development of NPM. For the typeset glyphs we use in this article we employ a design based on scripts of several natural languages.[3] We begin with the glyph for a set. If pressed, we will say this is the set concept determined formally by the Zermelo-Frankel axioms, including the axiom of choice:

We observe a vertical center line, to serve as a visual anchor, partially enclosed by a pair of dots (represented as they might be drawn using a brush). We can now begin developing the glyph as structural features are added to a set. This will give rise to the first of the two major families of glyphs, the *structure* glyphs. Following terminology for Chinese and Japanese, we call the most basic glyphs *radicals*, and thus we say that the set glyph is one of the basic *radicals* of the NPM system. From the set glyph immediately follow two of the most important radicals, that of groups and semigroups, and that of categories:

On the left is the NPM glyph of the group concept, and on the right is the glyph of the mathematical category concept. The depictive elements in these glyphs are briefly the following: a group is a set with a little bit of extra structure added (a binary operation satisfying various properties). This additional structure is represented by extending the dots of the set glyph to the ascending diagonal lines of the group glyph. Thus, via absence loading, when we write the glyph of a *set*, or *invoke* it, we simultaneously invoke all *groups*. (In other words, it is unspoken whether or not the sets we invoke are in fact groups.) On the other hand, the relationship between the category radical and the set radical is slightly irregular: a set is what is known as a *zero category*: in the language of category theory, this is a set of objects with no morphisms between them. There is an ascending sequence of 0-categories, 1-categories (which are the usual categories), 2-categories, 3-categories, etc. Since this sequence and the "higher category theory" it leads to is foundational to many areas of mathematics, we build into NPM the relationship between sets and categories. In other words, NPM views a set as a zero category—as well as a structure with a coherent definition and theoretical area of study of its own.

These characters can now be recombined to give the character for a *groupoid,* a category in which all morphisms are invertible:

Coincidentally, this character is reminiscent of the Han character for water, 氺. Guided by the Han character, we make the strokes slightly asymmetric, for this increases its aesthetic appeal and evokes the experience of writing the glyph by hand. There are numerous variants of these basic glyphs that will not be discussed in this overview.

Now we come to the second major family: the topological glyphs, which differ from the structure glyphs semantically, grammatically, as well as typographically. The topological family has two radicals instead of one, the *Kolmogorov space* radical (left) and the *Hausdorff space* radical (right):

[3]Elements of the design were influenced by, e.g., the Hiragana syllabary and Sanscrit alphabet, Arabic and Khmer scripts, and Chinese Han characters.

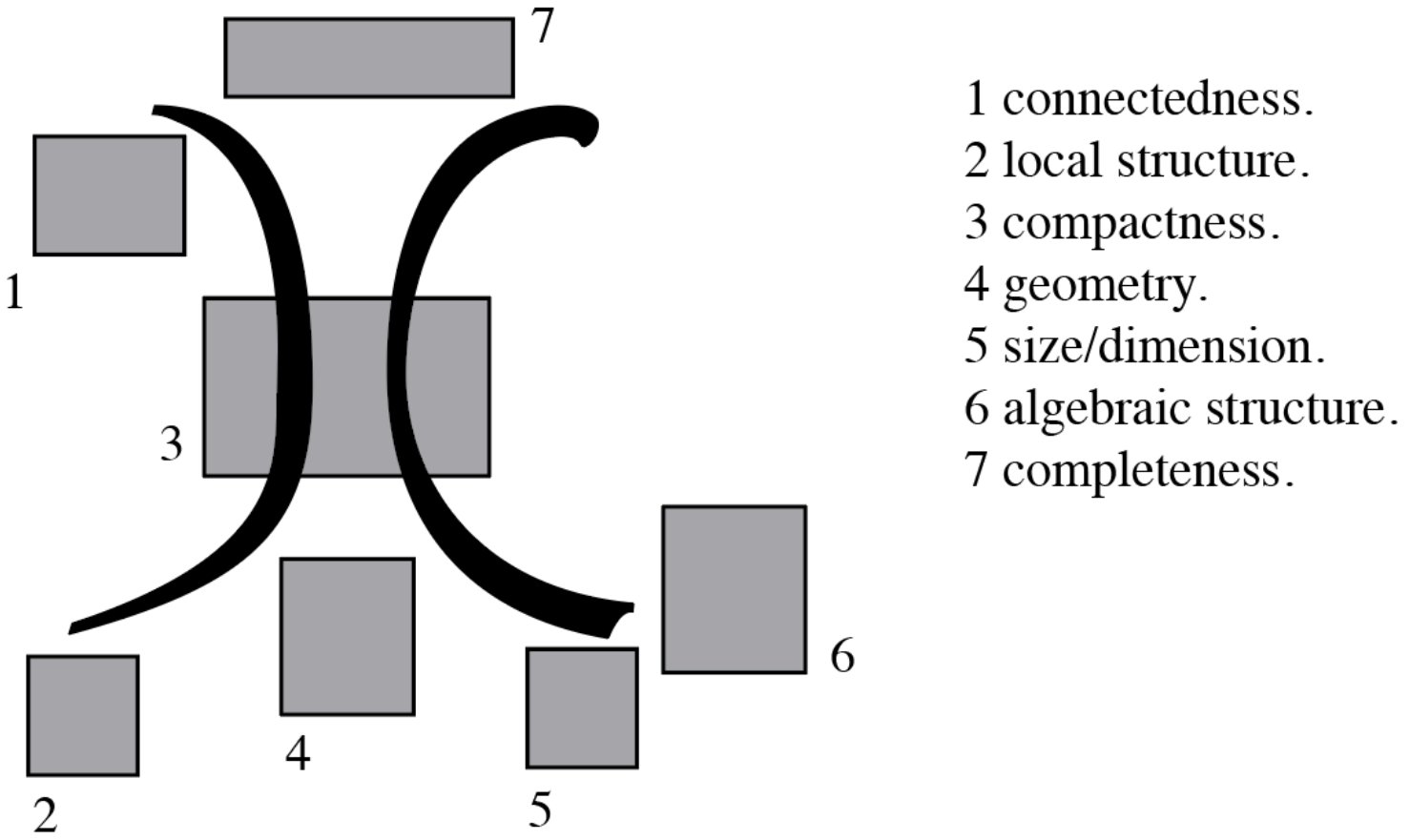

FIGURE 1. Absence-loaded regions of the Hausdorff space radical.

The treatment of the Hausdorff glyph as a radical reflects the importance of Hausdorff spaces, particularly in analytic areas of study (for example, harmonic analysis) where a topological space is needed that can support a notion of measure (in the sense of Lebesgue theory). Kolmogorov spaces, on the other hand, are often studied very differently. For example, they frequently arise in situations in which the so-called specialization ordering is an important part of the structure (the specialization ordering is trivial in the case of a Hausdorff space). These differences are reflected directly (depictively) in the associated NPM glyphs, in the sense that the Hausdorff property is not added to the Kolmogorov space radical, even though a priori this is how one might have expected to proceed.

For the sake of brevity, we will not develop the topological radical's many variants here. Figure 1 indicates absence-loaded regions of the radical, along with the class of properties reflected (or "loaded" or "held") in each region. Note that positions are disjoint and reflect design choices regarding proximity to the center of the glyph. In particular, compactness is placed directly in the center of the glyph: this reflects the general fact that compactness across virtually all of mathematics is among the most important and most impactful properties of a topological space. We note, in passing, that the glyph of a compact Hausdorff space is denoted in NPM by Here the intuition of compactness is represented depictively by a closed circle. It may also be noteworthy that the algebraic structure region's location is deliberately chosen to allow maximum flexibility concerning size adjustment (or "expansion" of a region), as a great deal of the variation in topological spaces arising in practice is due to variations in algebraic structure.

We will now give a demonstration of how NPM grammar patterns can be combined to create complex glyphs out of simple ones. For this we will study the development of structures arising in analysis. Let us go back to the group glyph

We can add the assumption that it is abelian by transforming the enclosing strokes into a cross:

This depictively evokes the symbol of addition, the Greek cross used by Widman (in 1489) and Descartes [Caj]. Next we can transform to a *vector space* by assuming the action of a field:

Now we begin to see that, in but a single glance, we can perceive via NPM the close relationship between many diverse structures, each with a distinct theory and semantic space of their own: vectors spaces, groups, sets, and categories.[4] Now we combine the topological and structure glyphs to generate the notion of a *topological vector space*:

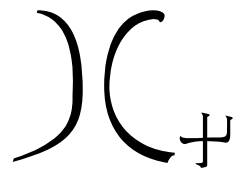

When such structures are studied, several properties come to the fore, generating a family of glyphs that in NPM take the form:

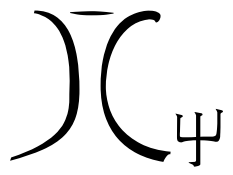 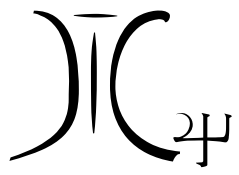 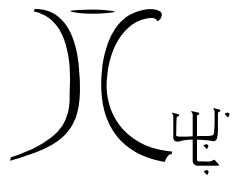

The additional strokes in the glyphs reflect structural and topological properties: *completeness*, the presence of an inner product structure, the presence of a norm structure, and several properties related to the presence of an involution [Dix]. These glyphs (at left, at center, at right) depict structures that are known in analysis as *Banach spaces*, *Hilbert spaces*, and $C^*$*-algebras*, respectively, each of which is the basis of a substantial mathematical theory.

Next, we illustrate how NPM provides representations that can concisely express otherwise obscure relationships between mathematical concepts. For this example, we return to the category radical, and we develop it to the notion of a topos, whose glyph is

[4]The concept of a *module* (in the mathematical sense) is also very closely related to the rest of these concepts, and is therefore represented in NPM by an evocatively similar glyph:

The line depicting the scalar action of a ring or algebra is not so far extended (ascends not as far as the cap height).

which is abbreviated,[5] for convenience, to There is more to tell, however, in unpacking this glyph. The terminology "topos" is in fact used to denote two mathematical concepts, the first known as *elementary topos* and the chronologically earlier concept of *Grothendieck topos*. The Grothendieck topos can be defined as an elementary topos that is cocomplete and has a small generating set. However, this definition does not reflect the (so to speak) "correct" intuition in all situations where the notion of a Grothendieck topos is used. Therefore the position of NPM is that a Grothendieck topos is denoted using a glyph based on a different radical than that of the elementary topos:

This glyph, which irregularly displays the category glyph in the geometric region (see Figure 1), evokes the Kolmogorov space radical above—in other words, it evokes a spatial object, not a structural one. This reflects the language of generalized spaces developed in the original work by Grothendieck and his school [SGA4], based on geometric morphisms. The fact that the two approaches to topos theory in fact converge on the same notion and a common underlying theory was the product of years of development by mathematicians; for better or for worse, NPM can capture the entire thrilling story, more or less, simply by invoking their names.

For the last demonstration, we show how NPM may be used to express logical relations between structures, and in particular its grammatical role with respect to morphisms. We choose a relation in order theory, known as *Priestley duality*:

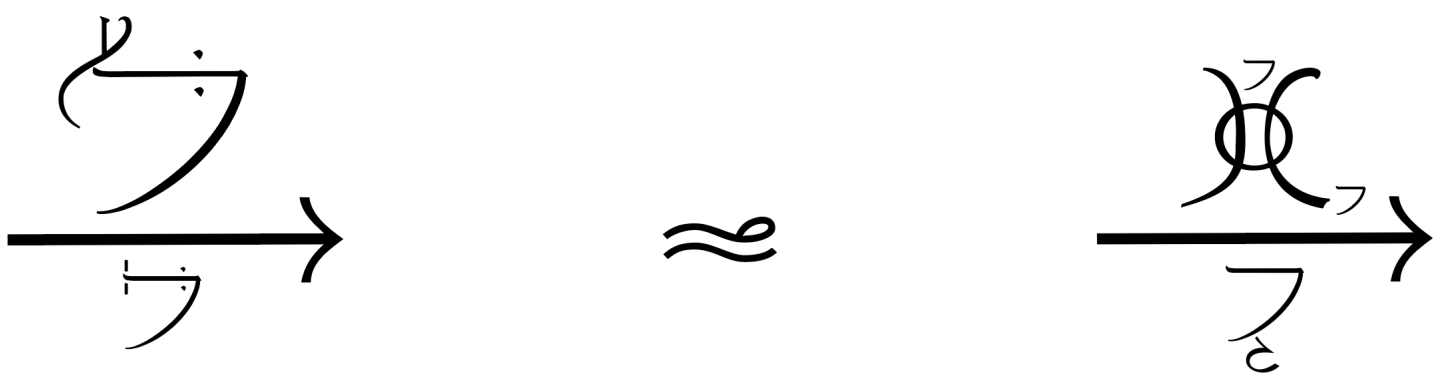

Here, a glyph set above an arrow denotes the objects of a category whose morphisms are described using the glyph centered below the arrow, and the symbol ≈ denotes a duality (i.e., an equivalence between one and the opposite of the other category). Within order theory there are at least two

[5]This abbreviation is a regular, recurring NPM pattern. The vertical line which is suppressed is called the *limit file*, and is decorated with absence-loaded regions for expressing the limit properties of a category. Thus a category glyph with limit file extended is written 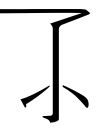, and extra strokes are added which, in the case above, for example, reflect the fact that the structure is assumed to be cartesian closed and to have all finite limits. This creates a depictive relationship between this and the order glyph 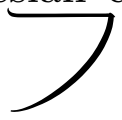 (see Table 1) that once again reflects underlying mathematical structure. Cf., for example,

which is the Heyting algebra glyph.

important cases of this duality: *Birkhoff duality*

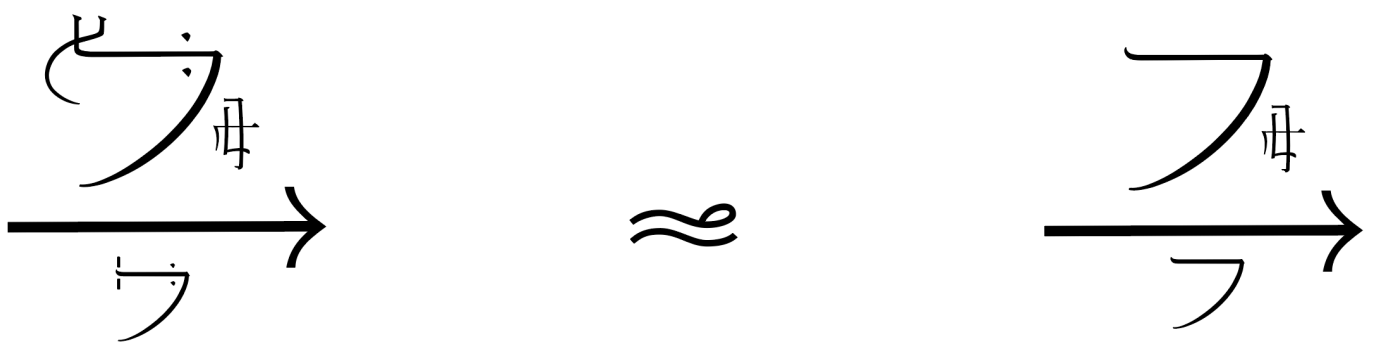

and *Stone duality*

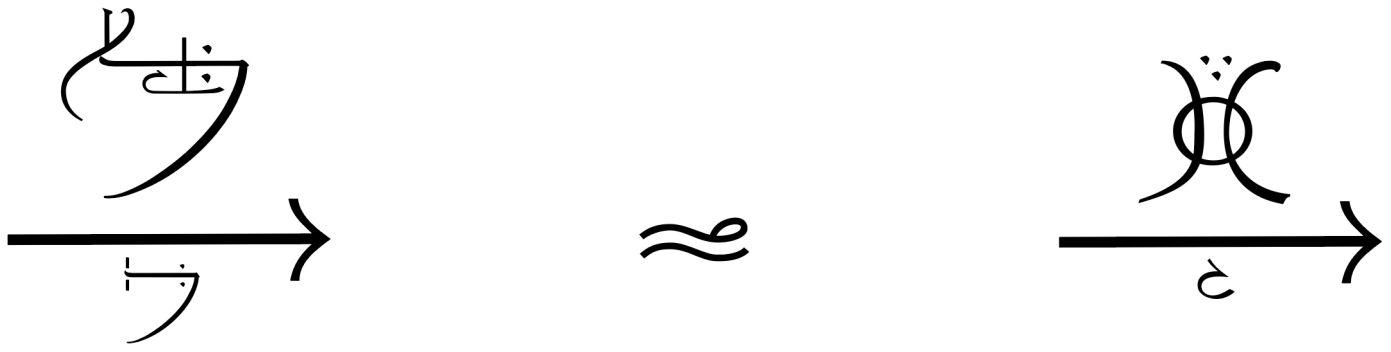

Notice that the NPM representations suggest, via highly automated cognitive processing mechanisms, that there are many relationships and patterns to be recognized between the three formulas. The task of bringing to the fore the same parallels in the structure of these statements cannot be as easily achieved via descriptive methods.[6] Variations of this broad duality relationship, which is a form of spectral duality, can be found not only in order theory, but in many other mathematical disciplines as well. NPM captures each one of these and often succeeds in capturing something of the metaphorical relationship between these different dualities in an evocative and information dense, calligraphic, yet useful form.

Finally, the entire collection of NPM radicals is presented in Table 1.

## 4. Conclusion and Outlook

The problem of developing an implementation of a visual type theory for mathematics such as NPM demands ventures in both the sciences and the arts, across the boundaries between mathematics, computer science, and graphic design, and poses many challenges. For its grounding as a tool for cognition, it appeals to fields such as educational and cognitive psychology for a model on which to base its mechanical function, and for empirical results that may clarify its role. The empirical results reviewed in section 2.1 are encouraging in this respect, but as these studies arise from experiments involving text adjuncts like (among others) graphs, charts, and representational images, we could regard this evidence as not directly hitting the bull's eye. For instance, the evidence on transformational and decorational text adjuncts in [LAC, DSL] suggests that an information dense, grammatically regular, depictive device like a visual type theory may in some settings boost learning and memory performance. However, absence loading (section 3.3), construed as a memnonic device and tool for schema-construction (cf. sections 2.5, 3.1, where it was noted that a visual type theory could be used as a public or as a private language for this purpose), still has not itself been scientifically tested or evaluated. It is also unclear how the prior knowledge demanded

[6] We cannot fully explain the glyphs appearing in the dualities here; however, 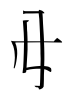 is the glyph of finiteness (the three strokes of the glyph correspond to ascending and descending chain condition, and cochain-finiteness, respectively), and 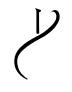 denotes distributivity. The rest can be inferred from formal statements.

| | | | |
|---|---|---|---|
| 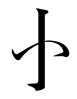 | set | 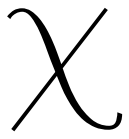 | sheaf (geometric view)* |
| 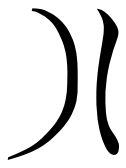 | Kolmogorov space | 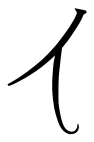 | (dynamical) system, triple |
| 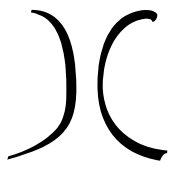 | Hausdorff space | 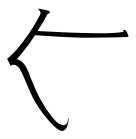 | process |
| 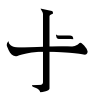 | ring, field, algebra | 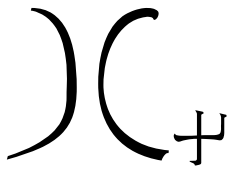 | topological vector space |
| 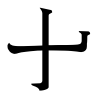 | module, vector space | 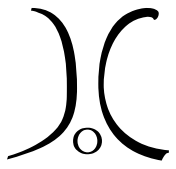 | CW complex |
| 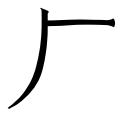 | pairs, extensions | 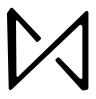 | simplicial set, Kan complex* |
| 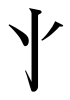 | group | 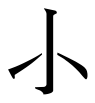 | category |
| 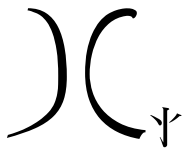 | group (topological) | 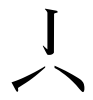 | globular set, generalized category** |
| 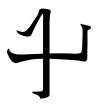 | Lie algebra | 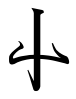 | enriched category |
| 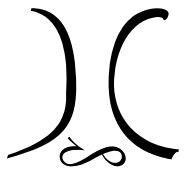 | manifold, bundle | 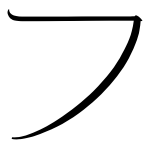 | order, lattice |
| 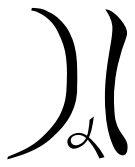 | classical variety | 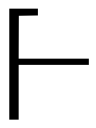 | deduction system, graph* |
| | | 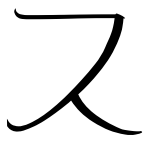 | lambda calculus |

* subject to revision
** recent/experimental

Table 1. NPM: The Basic Radicals, July 2016.

for the use of a visual type theory would affect its performance as a memnonic and organizing tool in practice. However, the author would regard it as too much to argue that the existing evidence falls off the target altogether, and would suggest that results so far are promising from the point of view of visual type theory, and that further research is warranted.

As regards potential NPM enthusiasts who are mathematically trained, there appears to be, fortunately, only two major axes along which NPM must be brought to smoother interoperability if NPM is to become usable: TeX platforms, since TeX is the near-universal language of mathematical and technical writing, and the space within and around the large family of proof assistants currently available (Coq, Agda, Isabelle, CompCert, Lean, and others), perhaps via adaptation of existing middleware such as the Proof General [Asp]. A first-generation NPM system could provide such inter-operability, however minimal. The creation and maintenance of a full library of NPM glyphs even on this minimal basis would require not only facility in typographical design, but also broad knowledge of mathematics, in order to maximize the quality and realism of the language. The author's hope is that the project is continued without losing the sense of purpose it has carried since its earliest beginnings: for the NPM developer, the task faced is not only to be of service to its users, but also to be a participant in the relationship, as old as civilization itself, between people and mathematics.

A working NPM platform would involve a usable glyph set easily numbering in the tens of millions. If such a platform is one day available, the problem of how to improve user interface can then be approached in conformity with a plan for growth. This suggests connections to the fields of mathematical user interface design and human-computer interaction, as well as programming language theory, which supplies an understanding of the anatomy and life cycles of formal languages existing in diverse and rapidly changing environments. Further study in these areas can perhaps provide models and guidelines which can be adapted to the visual type theory paradigm. All research related to NPM's development would also have to follow developments in a number of areas including proof assistants and proof engineering, mathematical typesetting, mathematical software, and mathematical knowledge management.

## 5. Acknowledgements

It is a privilege to do interdisciplinary research, and I deeply thank everyone who helped me put this still-incomplete puzzle together. I would especially like to thank the CICM 2016 and MathUI referees for their criticism and suggestions, which led to many significant improvements of this work.